\documentclass{midl} % Include author names

\usepackage{mwe} % to get dummy images

\usepackage{array, xcolor, colortbl}
\usepackage{multirow}
\usepackage{booktabs}
\usepackage{upgreek}
\usepackage{mathtools}
\usepackage{dsfont}
\usepackage{hyperref}

\definecolor{Gray}{gray}{0.85}
\definecolor{LightCyan}{rgb}{0.88,1,1}
\newcolumntype{F}{>{\columncolor{LightCyan}}c}
\newcolumntype{G}{>{\centering}p{0.05\textwidth}}

\jmlryear{2024}\jmlrworkshop{Full Paper -- MIDL 2024}\jmlrvolume{-- 64}\editors{Accepted for publication at MIDL 2024}

\title[Dense SSL for Segmentation]{Dense Self-Supervised Learning for Medical Image Segmentation}

 \midlauthor{\Name{Maxime Seince\nametag{$^{1,2}$}} \Email{maxime.seince22@imperial.ac.uk}\\
  \Name{Loïc {Le Folgoc}\nametag{$^{1}$}} \Email{loic.lefolgoc@telecom-paris.fr}\\
  \Name{Luiz Augusto {Facury de Souza}\nametag{$^{1}$}} \Email{luiz.facurydesouza@telecom-paris.fr}\\
  \Name{Elsa Angelini\nametag{$^{1}$}} \Email{elsa.angelini@telecom-paris.fr}\\
  \addr $^{1}$ LTCI, Telecom Paris, Institut Polytechnique de Paris, France\\
  \addr $^{2}$ Imperial College London, UK
  }

\begin{document}

\maketitle

\newcommand{\rev}[1]{\textcolor{black}{#1}}

\newcommand{\bh}{\mathbf{h}}
\newcommand{\bx}{\mathbf{x}}
\newcommand{\bu}{\mathbf{u}}
\newcommand{\bv}{\mathbf{v}}
\newcommand{\bz}{\mathbf{z}}
\newcommand{\un}{\mathds{1}}

\begin{abstract}
Deep learning has revolutionized medical image segmentation, but it relies heavily on high-quality annotations. The time, cost and expertise required to label images at the pixel-level for each new task has slowed down widespread adoption of the paradigm. We propose \texttt{Pix2Rep}, a self-supervised learning (SSL) approach for few-shot segmentation, that reduces the manual annotation burden by learning powerful pixel-level representations directly from unlabeled images. \texttt{Pix2Rep} is a novel pixel-level loss and pre-training paradigm for contrastive SSL on whole images. It is applied to generic encoder-decoder deep learning backbones (\textit{e.g.}, U-Net). Whereas most SSL methods enforce invariance of the learned \textit{image-level} representations under
intensity and spatial image augmentations, \texttt{Pix2Rep} enforces \textit{equi}variance of the \textit{pixel-level} representations. 
We demonstrate the framework on a task of cardiac MRI segmentation. Results show improved performance compared to existing semi- and self-supervised approaches; and a $5$-fold reduction in the annotation burden for equivalent performance versus a fully supervised U-Net baseline. This includes a $30\%$ (resp. $31\%$) DICE improvement for one-shot segmentation under linear-probing (resp. fine-tuning). Finally, we also integrate the novel \texttt{Pix2Rep} concept with the Barlow Twins non-contrastive SSL, which leads to even better segmentation performance.
\end{abstract}


\section{Introduction}

Medical image segmentation has seen tremendous progress with the advent of deep learning~\cite{Ronneberger2015,Milletari2016,Kamnitsas2017}. The drawback of this paradigm is its reliance on large quantities of data, annotated at the pixel-level, to train strong segmentation models. These pixel-level annotations are costly 
to obtain, and take precious time from medical experts. 

To circumvent this burden, techniques have emerged in recent years that better exploit more widely available \textit{un}labeled data. Semi-supervised approaches \textit{e.g.}, pseudo-labels~\cite{Lee2013,Bai2017,Tran2022} and mean teacher~\cite{Tarvainen2017,Yu2019}, balance a supervised segmentation loss on a small labeled dataset with a consistency loss on the larger unlabeled dataset, 
yielding improved segmentation. Other semi-supervised approaches include Bayesian deep learning \textit{e.g.},~\citet{Dalca2018} introduce anatomical priors that can be learnt using unlabeled or unpaired data.

Self-Supervised Learning (SSL) follows an alternative route whereby deep representations are directly learned from unlabeled data. Early methods trained these representations by solving pretext tasks \textit{e.g.}, relative position prediction~\cite{Doersch_2015_ICCV}, image recolorization~\cite{Zhang2016colorful}, jigsaw puzzles~\cite{Noroozi2016} or rotation prediction~\cite{gidaris2018unsupervised}. These methods are designed for image classification as a primary downstream task, thus an image is encoded to an image-level vector representation. Many recent methods for image-level representation learning coexist in the state-of-the-art, based on contrastive learning~\cite{chen20j,He_2020_CVPR}, on redundancy-reduction~\cite{zbontar21a}, on self-distillation~\cite{BYOL2020,Caron_2021_ICCV}, on Masked Image Modeling~\cite{He_2022_CVPR} and many more.

\textbf{We propose} instead \textbf{a framework for pixel-level (dense) representation learning}, dubbed \texttt{Pix2Rep}, which can be used to \textbf{pretrain encoder-decoder architectures}, such as U-Nets. Whereas most aforementioned methods rely on invariance under certain intensity-based augmentations (brightness \& contrast, Gaussian noise, etc.) and geometric augmentations (crops), \texttt{Pix2Rep} is based on \textit{equi}variance under geometric transformations. For the task of cardiac MRI segmentation, \textbf{we propose rotations and intensity reversals as additional augmentations} that further improve results. Finally, \textbf{we investigate the performance} of \texttt{Pix2Rep} \textbf{in various data regimes} (one-shot, few-shot segmentation, or large annotated data), both under linear probing and fine-tuning.

\section{Related Work}

Comparatively, fewer pixel-level representation learning methods have been proposed so far. \citet{Kalapos2022,Punn2022} propose to pretrain a U-Net encoder (a.k.a.~its downsampling branch) using image-level SSL (BYOL/Barlow Twins). The U-Net decoder however is randomly initialized before fine-tuning on the downstream segmentation task. \citet{Tang_2022_CVPR} pretrain a Swin UNETR encoder using a combination of image-level contrastive learning, pretext task and masked image modeling. \citet{Zeng2021} exploit the positional information of slices within stacks for contrastive pretraining of a U-Net encoder. 

\citet{NEURIPS2020_949686ec} manage pretraining of the first decoder layers by introducing a local contrastive loss that relies on rough alignment of subject volumes. For contrastive pretraining of the whole decoder~\cite{Xie_2021_CVPR}, positive pairs of \textit{pixels} need to be defined. 
\citet{Zhong_2021_ICCV} constrain the two augmented views to differ only up to intensity transformations, so as to form positive pairs from pixels at identical locations in the two views. \citet{Hu2021,Zhao_2021_ICCV} regard as positive all pixels sharing the same label, at the cost of pretraining only on the (potentially smaller amount of) labeled data. \citet{Wang_2021_CVPR,NEURIPS2022_39cee562} define pixels with highly similar features as positives. The closest related work to our proposed approach are those of \citet{NEURIPS2020_3000311c,Ke2022,Goncharov2023}, which define positive pairs to be pixels that describe the same physical location in the scene on different augmented patches, that differ up to intensity-based and geometric augmentations. 
Our framework 
targets equivariance rather than equivalence to random spatial transforms and works at whole image level for augmentations.
\citet{NEURIPS2020_3000311c} 
experiment on natural --not medical-- 
images. \citet{Goncharov2023,Ke2022} focus less on few-shot segmentation.

\section{Methods}

\begin{figure}
\floatconts
  {fig: pipeline}
  {\caption{
  Pretraining of arbitrary encoder-decoder architectures $f$ (\textit{e.g.}, U-Net). $\bx$ an unlabeled training image; $\phi\sim \mathcal{T}_s$ a random spatial transformation; $t,t'\sim\mathcal{T}_i$ two random intensity transformations; $g$ a projection head. We train pixel representation maps output by $f$ to be equivariant under $\phi$ and invariant to $t,t'$ by maximizing agreement between the outputs of the two branches, via a pixel-level contrastive loss.
  }
  }
  {\includegraphics[width=0.5\linewidth]{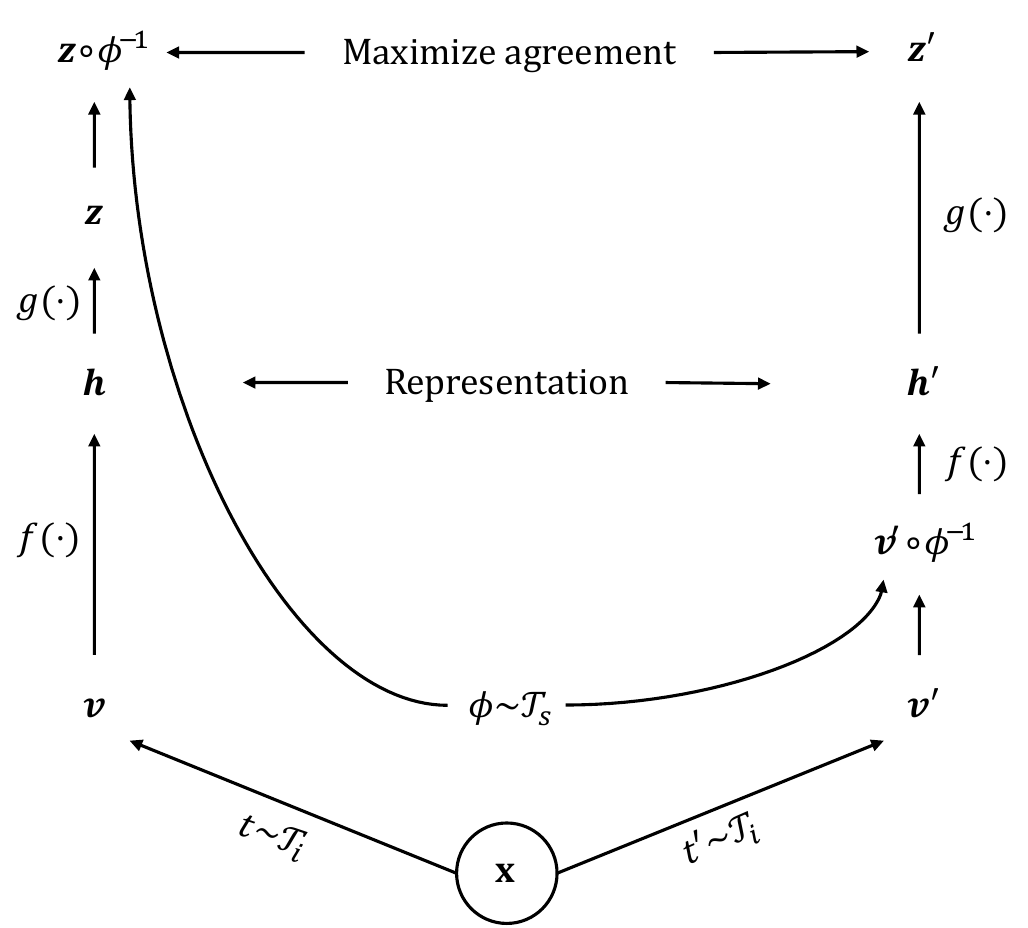}}
\end{figure}

We consider an arbitrary (trainable) neural network backbone $f:\mathbb{R}^{H\times W\times C}\mapsto \mathbb{R}^{H\times W\times D}$ that maps input images to pixel representation maps. $f$ can be any encoder-decoder network and we opt for simplicity for a U-Net~\cite{Ronneberger2015}, without its final segmentation head ($\equiv$ typically $1\times 1$ conv~+ softmax)\footnote{For downstream segmentation tasks, the segmentation head $\mathbb{R}^{H\times W\times D}\mapsto \mathbb{R}^{H\times W\times K}$, with $K$ the number of classes, is plugged back at the end of the pretrained backbone}. $f$ can be interpreted as embedding $C$-dimensional input pixels to $D$-dimensional vector representations, accounting for local and global context. At the core of the proposed self-supervised pretraining of $f$ is a contrastive loss which forces the pixel representations derived from $f$ to be invariant under the action of intensity augmentations and equivariant under the action of spatial transformations. The method is summarized in Fig.~\ref{fig: pipeline}.

\texttt{Pix2Rep} pretraining relies exclusively on an unlabeled dataset $\mathcal{D}\triangleq \{\bx\in \mathbb{R}^{H\times W\times C}\}$. Apart from the encoder-decoder $f(\cdot)$ which extracts pixel-level representation maps, the framework consists of three major components. 

(1)~For any input image $\bx$, a stochastic data augmentation module generates two random intensity-based transformations $t,t'\sim \mathcal{T}_i$ (incl.~brightness \& contrast augmentation, Gaussian noise, bias field, and intensity reversal), resulting in two views $\bv\triangleq t(\bx), \bv'\triangleq t'(\bx)$. In addition, the stochastic data augmentation module generates a single random spatial transformation $\phi\sim \mathcal{T}_s$ (incl. zooms, flips, rotations), which is applied asymmetrically to $\bv,\bv'$. 

(2)~A small projection head $g: \mathbb{R}^{H\times W\times D}\mapsto \mathbb{R}^{H\times W\times d}$ transfers pixel representation maps to the space where the contrastive loss is applied. The projection head $g(\cdot)\triangleq W^{(2)}\ast \sigma(W^{(1)}\ast \cdot)$ consists in a MLP with one hidden layer (where $\sigma$ is a ReLu non-linearity), implemented using $1\times 1$ convolutions. We have experimented with deeper projection heads (up to $3$ hidden layers) without noticing significant benefits. Then: 
\begin{itemize}
    \item For the view $\bv'$: we first transform it to the new spatial viewpoint by applying $\phi$, yielding\footnote{The action $\phi\cdot \bv'$ of a spatial transformation $\phi$ on an image $\bv'$ is $\bv'\circ \phi^{-1}$. This is well-known in the image registration literature. Following standard practice, we randomly sample $\phi^{-1}$ directly (rather than $\phi$) to circumvent the numerical inversion.} $\phi\cdot \bv'\triangleq \bv'\circ\phi^{-1} \in\mathbb{R}^{H\times W\times C}$, then pass the spatially-transformed image through $g\circ f$, yielding $\bz'\triangleq (g\circ f)(\phi\cdot \bv')$.
    \item For the view $\bv$: we first apply $g\circ f$, yielding the pixel representation map $\bz=(g\circ f)(\bv)$, before transporting $\bz$ to the new viewpoint: $\phi \cdot \bz = \phi \cdot (g\circ f)(\bv)$.
\end{itemize}
Note that $\phi$ is applied exactly once along the two computational branches, hence $\phi\cdot \bz=\phi\cdot (g\circ f)(t(\bx))$ and $\bz'=(g\circ f)(\phi\cdot t'(\bx)))$ share the same viewpoint.

(3)~Thirdly, a contrastive loss $\mathcal{L}$ 
is defined for a pixel-level contrastive prediction task. Given any pixel coordinate $s\in \mathbb{R}^2$, features $(\phi\cdot\bz)(s)\in \mathbb{R}^d$ and $\bz'(s)\in \mathbb{R}^d$ correspond to the same anatomical point in the two views, thus they form natural positive pairs. 
Natural negative pairs 
are obtained from samples from the same view at all other pixel coordinates, or from other images in the minibatch at any pixel coordinate. 
However, creating one positive pair per pixel coordinate would yield hundreds of thousands of positive pairs and potentially billions of negative pairs. 

Instead, we sample $M$ random pixel coordinates $\{s_m^{(n)}\}_{m=1\dots M}$ independently for each image $1\leq n\leq N_b$ in the minibatch. For a given image $\bx$ and a given coordinate $s$, we obtain a positive pair of examples $\bu_i\triangleq(\phi\cdot\bz)(s)$, $\bu_j\triangleq \bz'(s)$. The negative examples for this positive pair $(\bu_i,\bu_j)$ are obtained from the other 
$N_bM-1$ pairs of positive examples across all sampling coordinates and images in the minibatch. In total, this generates $N\triangleq N_bM$ positive pairs and $2(N_bM-1)$ negative examples for each positive pair. By analogy to SimCLR~\cite{chen20j}, we use the InfoNCE loss $l_{i,j}$ of Eq.~\eqref{eq: InfoNCE positive pair} for each positive pair $(\bu_i,\bu_j)$:

\begin{equation}
    l_{i,j}\triangleq -\log{\frac{\exp\left(\text{sim}(\bu_i,\bu_j)/\tau\right)}{\sum_{k=1}^{2N}{\un[k\neq i]\exp\left(\text{sim}(\bu_i,\bu_k)/\tau\right)}}}\ ,
    \label{eq: InfoNCE positive pair}
\end{equation}
where $\tau$ denotes a temperature parameter, and $\text{sim}(\bu,\bu')\triangleq \bu^T\bu'/\Vert\bu\Vert_2\Vert\bu'\Vert_2$ denotes the cosine similarity. The total loss is aggregated by summing over all $l_{i,j}$, including symmetrizing the roles of $i,j$. Assuming 
without loss of generality that positive pairs have consecutive indices $i=2k-1$ and $j=2k$ in the list of sampled pixels, this yields Eq.~\eqref{eq: total loss}:

\begin{equation}
    \mathcal{L}\triangleq \frac{1}{2N}\sum_{k=1}^N \left(l_{2k-1,2k}+l_{2k,2k-1}\right)\ .
    \label{eq: total loss}
\end{equation}

\textbf{Pix2Rep-v2.} The computational complexity of contrasting positive and negative pairs of pixels limits us to sample $M$ 
coordinates for the InfoNCE contrastive loss. Alternatively, we propose to replace this contrastive loss with a loss based on \texttt{Barlow Twins}~\cite{zbontar21a}: we call this variant \texttt{Pix2Rep-v2}. 
It minimizes the \texttt{Barlow Twins} loss defined from the cross-correlation matrix $\mathcal{C}\in\mathbb{R}^{d\times d}$ between twin pixel embeddings $(\phi\cdot\bz)(s)$ and $\bz'(s)\in\mathbb{R}^d$, aggregated over all pixel coordinates $s$ and the whole minibatch. Although it aggregates information from the whole pixel representation maps (rather than samples), \texttt{Pix2Rep-v2} has a reduced memory footprint compared to \texttt{Pix2Rep}'s contrastive loss.\\

\textbf{Downstream segmentation.} For a given segmentation task, we initialize the encoder-decoder $f$ with the pretrained weights (discarding the projection head $g$), and add a task-specific, learnable segmentation head ($1\times 1$ conv + softmax), projecting pixel representations to class probabilities. We keep $f$ frozen, and only train the segmentation head (called \textbf{linear probing}), or allow $f$ to be fine-tuned from the supervised data (called \textbf{fine-tuning}). 

\section{Experiments}

We demonstrate 
the self-supervised pretraining on a downstream task of cardiac MRI segmentation.\\ 

\begin{figure}
\floatconts
  {fig: similarity}
  {\caption{Pixel embedding similarity maps. Large images: query images in which we select a query pixel (highlighted in red). For each query, we display two test images, with the pixel closest (in embedding space) to the query pixel highlighted in red. Similarity maps (cosine similarity between pixel embeddings) are also shown.}}
  {\includegraphics[width=1\textwidth]{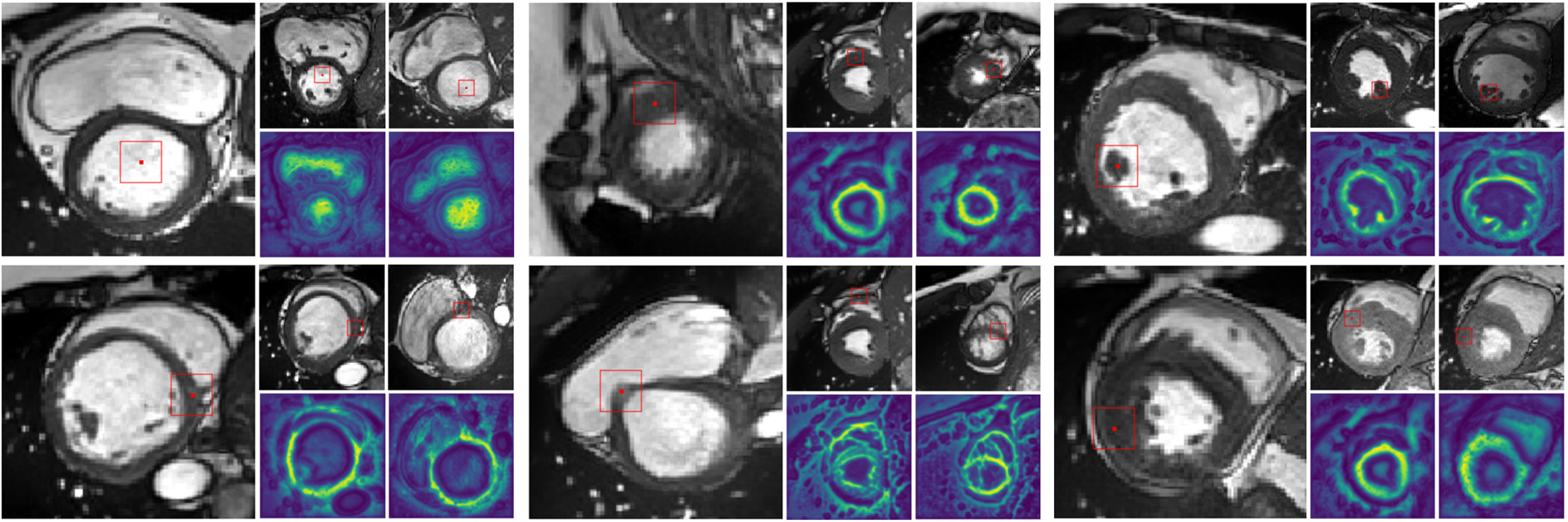}}
\end{figure}

\noindent\textbf{Data.} The ACDC dataset~\cite{Bernard2018} consists of $3D$ short-axis cardiac cine MR images of $150$ subjects, including 
expert annotations at End-Systole 
and End-Diastole 
for 
the left ventricle, right ventricle and myocardium. 
It is split into a training-validation set ($100$ images) and a test set ($50$ images). Slices are intensity-normalized using min-max normalization (using the 1st and 99th percentiles), cropped and resized to $128\times 128$.\\

\noindent\textbf{Experimental setup and Evaluation.} We use the provided split between training data and test data. The test data ($|X_{ts}|=50$) is only used for the final evaluation. For pretraining (via \texttt{Pix2Rep} or other approaches), we use the entirety of the raw training data ($|X_{pre}|=100$), without segmentation labels. For linear probing, fine-tuning, or fully-supervised training, we reuse a smaller number of training images with their segmentation labels ($|X_{tr}|\in\{1,2,5,10,20,50,100\}$). The slices in the stacks of $X_{tr}$ are divided between training data ($90\%$) and validation data ($10\%$).

We quantify the contribution of the \texttt{Pix2Rep} pretraining on the performance on a downstream segmentation task. The $3D$ Dice similarity coefficient is used as the evaluation metric. We report the average Dice over the test set over the segmented structures. For each evaluated method, each reported score is an average over five runs (training+test).\\

\begin{figure}
\floatconts
  {fig: vs. fully-supervised}
  {\caption{Proposed pretraining vs. fully-supervised baseline (same U-Net architecture).}}
  {\includegraphics[width=0.75\textwidth]{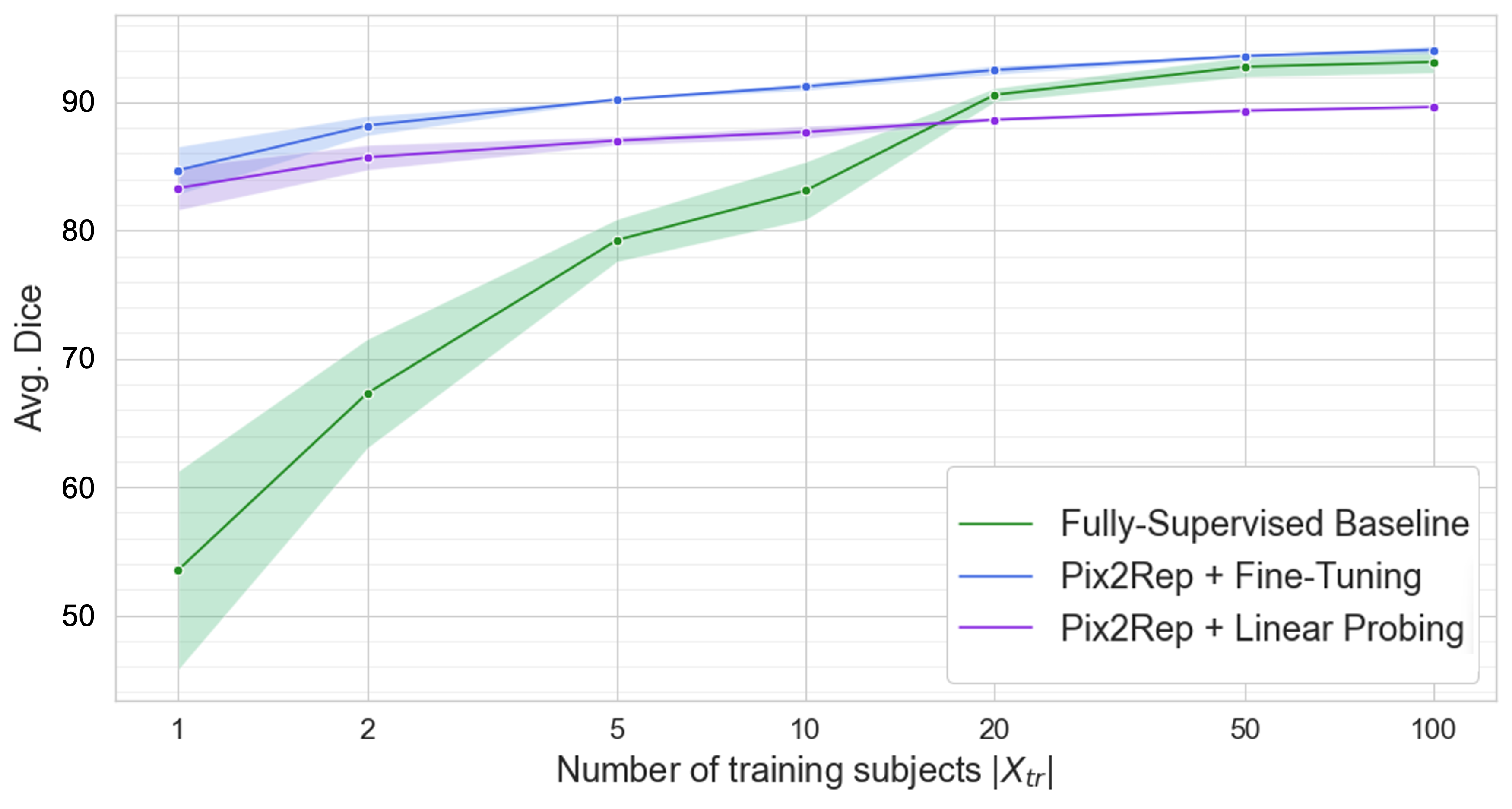}}
\end{figure}

\noindent\textbf{Comparison to other methods.} 
A natural baseline is obtained by keeping the same backbone 
U-Net, initialized with random (rather than pretrained) weights, then trained using varying amounts of labeled data $X_{tr}$ ($|X_{tr}|\in\{1,2,5,10,20,50,100\}$). We refer to it as the fully-supervised baseline. 

In addition, we compare with several methods in the state-of-the-art. We implement a mean teacher~\cite{Tarvainen2017} semi-supervised model (using the same U-Net backbone), where the supervised loss applies to the available labeled slices, and the unsupervised consistency loss uses all slices from the $100$ raw training volumes. As for self-supervised models, we compare to (1)~encoder-only pretraining of the same U-Net backbone, using image-level SimCLR, representative of~\citet{Kalapos2022,Punn2022}; (2)~encoder-decoder (\texttt{Pix2Rep}) pretraining without rotations (crops only) and without intensity reversals as augmentations; and (3)~the method of \citet{NEURIPS2020_949686ec}. Unless stating otherwise, all self-supervised approaches use fine-tuning in the second stage rather than linear probing, as this yields higher performance. For the proposed approach, we report numbers both for linear-probing and fine-tuning.\\

\noindent\textbf{Backbone and Implementation details.} We use a standard U-Net architecture consisting of five \texttt{DoubleConv} encoding blocks ({\texttt{\{Conv,BatchNorm,ReLu,Conv,BatchNorm,ReLu\}}}) with \texttt{MaxPooling} downsampling and five \texttt{DoubleConv} decoding blocks with \texttt{Trans-Conv} upsampling, starting with $128$ feature maps and doubling after each block up to $2048$ at the bottleneck. We experimented with changing the number of feature maps in the last layer (cf.~ablation study), starting at $64$, noting a monotonic improvement in performance up to the maximum tested value of $1024$. We report the performance for $n_{ft}\coloneqq 1024$. For the fully-supervised baseline instead, we noticed decreased performance above and below $n_{ft}\coloneqq 128$ and report performance for this setting.

The framework is implemented in \texttt{PyTorch} (\url{https://github.com/pix2rep/}). The spatial transformation $\phi$ is applied in auto-differentiable manner via \texttt{affine\_grid}, \texttt{grid\_sample}. Intensity reversal augmentations correspond to $\tilde{\bx}\coloneqq 1-\bx$. We pretrain using the proposed approach for $200$ epochs with the \texttt{Adam} optimizer. The batch size is set to {$8$}, the number of sampled pixels to {$1000$}. The learning rate for the backbone is set to {$5\cdot 10^{-4}$}. For fine-tuning ($100$ epochs), the learning rate of the backbone is set to {$5\cdot 10^{-5}$} and of the classification head to {$10^{-2}$}. For linear probing ($100$ epochs), the learning rate is set to {$10^{-2}$}.\\

\noindent\textbf{Results.} To gain qualitative insight into the learnt pixel representations, Fig.~\ref{fig: similarity} plots the cosine similarity between a query pixel embedding and other pixel embeddings in two other representative images. The semantics of the anatomical structures are captured to some extent, as pixels anatomically similar to the query pixel have the highest similarity with it.

\begin{table}[h]
    \centering
    \begin{tabular}{p{0.42\textwidth} p{0.05\textwidth} p{0.05\textwidth} p{0.05\textwidth} p{0.05\textwidth} p{0.05\textwidth} p{0.05\textwidth} p{0.05\textwidth}}
            & \multicolumn{7}{c}{\cellcolor{LightCyan}\textbf{Avg.~Dice (ACDC), for $|X_{tr}|$ set to:}} \\
    \bottomrule
        \multicolumn{8}{c}{\vspace{-0.85em}} \\
        \textbf{Method:\hspace{1.0cm}~/\hspace{1.0cm}~$|X_{tr}|\coloneqq$} & $1$ & $2$ & $5$ & $10$ & $20$ & $50$ & $100$ \\
    \bottomrule
        \rowcolor{Gray}
        \multicolumn{8}{c}{\textbf{Fully-supervised learning:}} \\
        \multicolumn{8}{c}{\vspace{-0.95em}} \\
        Baseline (U-Net) & $53.5$ & $67.3$ & $79.2$ & $83.1$ & $90.6$ & $92.8$ & $93.1$ \\
        \rowcolor{Gray}
        \multicolumn{8}{c}{\textbf{Semi-supervised learning:}} \\
        \multicolumn{8}{c}{\vspace{-0.95em}} \\
        Mean Teacher & $57.3$ & $68.9$ & $84.5$ & $89.0$ & $90.3$ & $92.4$ & $93.9$ \\
        \rowcolor{Gray}
        \multicolumn{8}{c}{\textbf{Self-supervised learning (+ Fine-tuning by default)}:} \\
        \multicolumn{8}{c}{\vspace{-0.95em}} \\
        SimCLR pretraining & $63.3$ & $77.8$ & $85.0$ & $88.9$ & $90.2$ & $92.1$ & $92.6$ \\
        Chaitanya et al. (2020) & $76.7$ & $78.0$ & $85.9$ & $88.7$ & $90.8$ & $92.2$ & $92.7$ \\
        \textbf{Proposed} (w/o rot. \& int. reversal) & $77.7$ & $80.4$ & $87.9$ & $90.2$ & $92.2$ & $93.3$ & $\mathbf{94.3}$ \\
        \textbf{Proposed} (only Linear-probing) & $83.3$ & $85.7$ & $87.0$ & $87.7$ & $88.6$ & $89.3$ & $89.6$ \\
        \textbf{Proposed} & \underline{$84.7$} & \underline{$88.2$} & \underline{$90.2$} & \underline{$91.2$} & \underline{$92.5$} & \underline{$93.6$} & $94.1$ \\
        \rowcolor{Gray}
        \multicolumn{8}{c}{\textbf{Combined method:}} \\
        \multicolumn{8}{c}{\vspace{-0.95em}} \\
        \textbf{Proposed} + Mean Teacher & $\mathbf{86.1}$ & $\mathbf{89.7}$ & $\mathbf{91.1}$ & $\mathbf{91.9}$ & $\mathbf{92.7}$ & $\mathbf{93.8}$ & \underline{$94.2$} \\
    \bottomrule
    \end{tabular}
    \caption{Comparison with the state-of-the-art on the test ACDC dataset (avg.~Dice score), for various amounts of labeled training data $|X_{tr}|$. Best results in \textbf{bold} and second best \underline{underlined}. Each score is averaged over five runs (see main text -- Evaluation).}
    \label{tab: benchmark}
\end{table}

\tableref{tab: benchmark} summarizes the main results. The proposed dense contrastive pretraining yields higher Dice than the fully-supervised baseline for all data regimes, as also shown in Fig.~\ref{fig: vs. fully-supervised}. The gap is largest for few-shot segmentation; 
there is a $29.8\%$ (resp.~$31.2\%$) improvement in Dice for one-shot segmentation in linear probing (resp.~fine-tuning). Of note is the limited drop in performance with linear probing even in the large data regime, highlighting the quality of the unsupervised nonlinear pixel representations. 
With fine-tuning, there is roughly a $5\times$ reduction in the annotation burden for equivalent performance vs.~the fully supervised baseline. 
Pretraining also reduces the variability in performance resulting from the choice of training subject, as shown by lower standard deviations in Fig.~\ref{fig: vs. fully-supervised}.\\

\noindent\textbf{Ablation study.} \tableref{tab: ablation study} shows the incremental contributions of various modifications on the proposed framework, starting from a suboptimal setup, in the few-shot segmentation setting (for $|X_{tr}|=5$). The inclusion of image intensity reversals (\textit{i.e.}, $\tilde{\bx}\coloneqq 1-\bx$) and rotations as augmentations has a significant impact on performance.

\begin{table}[h]
    \centering
    \begin{tabular}{p{0.5\textwidth} F}
    \toprule
        \textbf{\texttt{Pix2Rep} pretraining settings:} & \textbf{Avg.~Dice} \\
    \bottomrule
        \vspace{-0.75em} & \vspace{-0.75em} \\
        Defaults w/ {$10$} epochs pretraining & $39.7 \pm 1.0$ \\
        {$10$} epochs $\rightarrow 100$ epochs & $51.2 \pm 2.7$ \\
        $64$ feature maps $\rightarrow 1024$ feature maps & $73.4 \pm 0.9$ \\
        Add `intensity reversal' augmentation & $82.5 \pm 3.0$ \\
        Add rotation augmentation & $84.5 \pm 1.0$ \\
        Linear-probing $\rightarrow$ Fine-tuning & $89.2 \pm 1.3$ \\
        
    \bottomrule
    \end{tabular}
    \caption{Ablation study ($|X_{tr}|=5$). Self-supervised pretraining benefits from the increased number of feature maps as well as from the additional augmentations.}
    \label{tab: ablation study}
\end{table}

\noindent\textbf{Pix2Rep-v2.} \tableref{tab: Pix2Rep-v2} reports results obtained with the non-contrastive variant \texttt{Pix2Rep-v2}, which generally improves slightly over \texttt{Pix2Rep}.

\begin{table}[h]
    \centering
    \begin{tabular}{p{0.3\textwidth} p{0.05\textwidth} p{0.05\textwidth} p{0.05\textwidth} p{0.05\textwidth} p{0.05\textwidth} p{0.05\textwidth} p{0.05\textwidth}}
    \bottomrule
        \textbf{Method:\hspace{0.5cm}~/\hspace{0.5cm}~$|X_{tr}|\coloneqq$} & $1$ & $2$ & $5$ & $10$ & $20$ & $50$ & $100$ \\
    \bottomrule
        w/ Linear probing & $82.6$ & $85.0$ & $\mathbf{88.1}$ & $\mathbf{89.1}$ & $\mathbf{89.8}$ & $\mathbf{90.4}$ & $\mathbf{90.7}$ \\
        w/ Fine-tuning & $\mathbf{85.6}$ & $\mathbf{88.4}$ & $\mathbf{91.1}$ & $\mathbf{92.0}$ & $\mathbf{92.6}$ & $93.4$ & $94.0$ \\
    \bottomrule
    \end{tabular}
    \caption{Avg.~Dice score (ACDC dataset), for the proposed \texttt{Pix2Rep-v2} approach. Numbers in \textbf{bold} when above their \texttt{Pix2Rep} counterpart.
    }
    \label{tab: Pix2Rep-v2}
\end{table}

\vspace{-0.7cm}

\section{Discussion \& Conclusion}

We have introduced \texttt{Pix2Rep}, a novel framework for pixel-level (dense) self-supervised representation learning, that allows to pretrain encoder-decoder architectures such as U-Nets directly from unlabeled images. We have shown performance gains on a downstream cardiac MRI segmentation task. Especially in the few-shot segmentation regime for the most challenging structure, we got $83\%$ Dice with 5 training subjects vs. $70\%$ for the fully-supervised baseline, and closest SOTA method $3\%$ below. As future work, in addition to comparing to \citet{Ke2022}, we plan to evaluate the framework on various segmentation tasks and assess the generalizability of learned representations for novel tasks (foundation model).

\bibliography{midl24_64}

\begin{thebibliography}{34}
\providecommand{\natexlab}[1]{#1}
\providecommand{\url}[1]{\texttt{#1}}
\expandafter\ifx\csname urlstyle\endcsname\relax
  \providecommand{\doi}[1]{doi: #1}\else
  \providecommand{\doi}{doi: \begingroup \urlstyle{rm}\Url}\fi

\bibitem[Bai et~al.(2017)Bai, Oktay, Sinclair, Suzuki, Rajchl, Tarroni, Glocker, King, Matthews, and Rueckert]{Bai2017}
Wenjia Bai, Ozan Oktay, Matthew Sinclair, Hideaki Suzuki, Martin Rajchl, Giacomo Tarroni, Ben Glocker, Andrew King, Paul~M. Matthews, and Daniel Rueckert.
\newblock Semi-supervised learning for network-based cardiac mr image segmentation.
\newblock In Maxime Descoteaux, Lena Maier-Hein, Alfred Franz, Pierre Jannin, D.~Louis Collins, and Simon Duchesne, editors, \emph{Medical Image Computing and Computer-Assisted Intervention -- MICCAI 2017}, pages 253--260, Cham, 2017. Springer International Publishing.
\newblock ISBN 978-3-319-66185-8.

\bibitem[Bardes et~al.(2022)Bardes, Ponce, and LeCun]{NEURIPS2022_39cee562}
Adrien Bardes, Jean Ponce, and Yann LeCun.
\newblock Vicregl: Self-supervised learning of local visual features.
\newblock In S.~Koyejo, S.~Mohamed, A.~Agarwal, D.~Belgrave, K.~Cho, and A.~Oh, editors, \emph{Advances in Neural Information Processing Systems}, volume~35, pages 8799--8810. Curran Associates, Inc., 2022.
\newblock URL \url{https://proceedings.neurips.cc/paper_files/paper/2022/file/39cee562b91611c16ac0b100f0bc1ea1-Paper-Conference.pdf}.

\bibitem[Bernard et~al.(2018)Bernard, Lalande, Zotti, Cervenansky, Yang, Heng, Cetin, Lekadir, Camara, Gonzalez~Ballester, Sanroma, Napel, Petersen, Tziritas, Grinias, Khened, Kollerathu, Krishnamurthi, Rohé, Pennec, Sermesant, Isensee, Jäger, Maier-Hein, Full, Wolf, Engelhardt, Baumgartner, Koch, Wolterink, Išgum, Jang, Hong, Patravali, Jain, Humbert, and Jodoin]{Bernard2018}
Olivier Bernard, Alain Lalande, Clement Zotti, Frederick Cervenansky, Xin Yang, Pheng-Ann Heng, Irem Cetin, Karim Lekadir, Oscar Camara, Miguel~Angel Gonzalez~Ballester, Gerard Sanroma, Sandy Napel, Steffen Petersen, Georgios Tziritas, Elias Grinias, Mahendra Khened, Varghese~Alex Kollerathu, Ganapathy Krishnamurthi, Marc-Michel Rohé, Xavier Pennec, Maxime Sermesant, Fabian Isensee, Paul Jäger, Klaus~H. Maier-Hein, Peter~M. Full, Ivo Wolf, Sandy Engelhardt, Christian~F. Baumgartner, Lisa~M. Koch, Jelmer~M. Wolterink, Ivana Išgum, Yeonggul Jang, Yoonmi Hong, Jay Patravali, Shubham Jain, Olivier Humbert, and Pierre-Marc Jodoin.
\newblock Deep learning techniques for automatic mri cardiac multi-structures segmentation and diagnosis: Is the problem solved?
\newblock \emph{IEEE Transactions on Medical Imaging}, 37\penalty0 (11):\penalty0 2514--2525, 2018.
\newblock \doi{10.1109/TMI.2018.2837502}.

\bibitem[Caron et~al.(2021)Caron, Touvron, Misra, J\'egou, Mairal, Bojanowski, and Joulin]{Caron_2021_ICCV}
Mathilde Caron, Hugo Touvron, Ishan Misra, Herv\'e J\'egou, Julien Mairal, Piotr Bojanowski, and Armand Joulin.
\newblock Emerging properties in self-supervised vision transformers.
\newblock In \emph{Proceedings of the IEEE/CVF International Conference on Computer Vision (ICCV)}, pages 9650--9660, October 2021.

\bibitem[Chaitanya et~al.(2020)Chaitanya, Erdil, Karani, and Konukoglu]{NEURIPS2020_949686ec}
Krishna Chaitanya, Ertunc Erdil, Neerav Karani, and Ender Konukoglu.
\newblock Contrastive learning of global and local features for medical image segmentation with limited annotations.
\newblock In H.~Larochelle, M.~Ranzato, R.~Hadsell, M.F. Balcan, and H.~Lin, editors, \emph{Advances in Neural Information Processing Systems}, volume~33, pages 12546--12558. Curran Associates, Inc., 2020.
\newblock URL \url{https://proceedings.neurips.cc/paper_files/paper/2020/file/949686ecef4ee20a62d16b4a2d7ccca3-Paper.pdf}.

\bibitem[Chen et~al.(2020)Chen, Kornblith, Norouzi, and Hinton]{chen20j}
Ting Chen, Simon Kornblith, Mohammad Norouzi, and Geoffrey Hinton.
\newblock A simple framework for contrastive learning of visual representations.
\newblock In Hal~Daumé III and Aarti Singh, editors, \emph{Proceedings of the 37th International Conference on Machine Learning}, volume 119 of \emph{Proceedings of Machine Learning Research}, pages 1597--1607. PMLR, 13--18 Jul 2020.
\newblock URL \url{https://proceedings.mlr.press/v119/chen20j.html}.

\bibitem[Dalca et~al.(2018)Dalca, Guttag, and Sabuncu]{Dalca2018}
Adrian~V. Dalca, John Guttag, and Mert~R. Sabuncu.
\newblock Anatomical priors in convolutional networks for unsupervised biomedical segmentation.
\newblock In \emph{Proceedings of the IEEE Conference on Computer Vision and Pattern Recognition (CVPR)}, June 2018.

\bibitem[Doersch et~al.(2015)Doersch, Gupta, and Efros]{Doersch_2015_ICCV}
Carl Doersch, Abhinav Gupta, and Alexei~A. Efros.
\newblock Unsupervised visual representation learning by context prediction.
\newblock In \emph{Proceedings of the IEEE International Conference on Computer Vision (ICCV)}, December 2015.

\bibitem[Gidaris et~al.(2018)Gidaris, Singh, and Komodakis]{gidaris2018unsupervised}
Spyros Gidaris, Praveer Singh, and Nikos Komodakis.
\newblock Unsupervised representation learning by predicting image rotations, 2018.

\bibitem[Goncharov et~al.(2023)Goncharov, Soboleva, Kurmukov, Pisov, and Belyaev]{Goncharov2023}
Mikhail Goncharov, Vera Soboleva, Anvar Kurmukov, Maxim Pisov, and Mikhail Belyaev.
\newblock vox2vec: A framework for self-supervised contrastive learning of voxel-level representations in medical images.
\newblock In Hayit Greenspan, Anant Madabhushi, Parvin Mousavi, Septimiu Salcudean, James Duncan, Tanveer Syeda-Mahmood, and Russell Taylor, editors, \emph{Medical Image Computing and Computer Assisted Intervention -- MICCAI 2023}, pages 605--614, Cham, 2023. Springer Nature Switzerland.
\newblock ISBN 978-3-031-43907-0.

\bibitem[Grill et~al.(2020)Grill, Strub, Altch\'{e}, Tallec, Richemond, Buchatskaya, Doersch, Avila~Pires, Guo, Gheshlaghi~Azar, Piot, kavukcuoglu, Munos, and Valko]{BYOL2020}
Jean-Bastien Grill, Florian Strub, Florent Altch\'{e}, Corentin Tallec, Pierre Richemond, Elena Buchatskaya, Carl Doersch, Bernardo Avila~Pires, Zhaohan Guo, Mohammad Gheshlaghi~Azar, Bilal Piot, koray kavukcuoglu, Remi Munos, and Michal Valko.
\newblock Bootstrap your own latent - a new approach to self-supervised learning.
\newblock In H.~Larochelle, M.~Ranzato, R.~Hadsell, M.F. Balcan, and H.~Lin, editors, \emph{Advances in Neural Information Processing Systems}, volume~33, pages 21271--21284. Curran Associates, Inc., 2020.
\newblock URL \url{https://proceedings.neurips.cc/paper_files/paper/2020/file/f3ada80d5c4ee70142b17b8192b2958e-Paper.pdf}.

\bibitem[He et~al.(2020)He, Fan, Wu, Xie, and Girshick]{He_2020_CVPR}
Kaiming He, Haoqi Fan, Yuxin Wu, Saining Xie, and Ross Girshick.
\newblock Momentum contrast for unsupervised visual representation learning.
\newblock In \emph{Proceedings of the IEEE/CVF Conference on Computer Vision and Pattern Recognition (CVPR)}, June 2020.

\bibitem[He et~al.(2022)He, Chen, Xie, Li, Doll\'ar, and Girshick]{He_2022_CVPR}
Kaiming He, Xinlei Chen, Saining Xie, Yanghao Li, Piotr Doll\'ar, and Ross Girshick.
\newblock Masked autoencoders are scalable vision learners.
\newblock In \emph{Proceedings of the IEEE/CVF Conference on Computer Vision and Pattern Recognition (CVPR)}, pages 16000--16009, June 2022.

\bibitem[Hu et~al.(2021)Hu, Zeng, Xu, and Shi]{Hu2021}
Xinrong Hu, Dewen Zeng, Xiaowei Xu, and Yiyu Shi.
\newblock Semi-supervised contrastive learning for label-efficient medical image segmentation.
\newblock In Marleen de~Bruijne, Philippe~C. Cattin, St{\'e}phane Cotin, Nicolas Padoy, Stefanie Speidel, Yefeng Zheng, and Caroline Essert, editors, \emph{Medical Image Computing and Computer Assisted Intervention -- MICCAI 2021}, pages 481--490, Cham, 2021. Springer International Publishing.
\newblock ISBN 978-3-030-87196-3.

\bibitem[Kalapos and Gyires-T{\'o}th(2023)]{Kalapos2022}
Andr{\'a}s Kalapos and B{\'a}lint Gyires-T{\'o}th.
\newblock Self-supervised pretraining for 2d medical image segmentation.
\newblock In Leonid Karlinsky, Tomer Michaeli, and Ko~Nishino, editors, \emph{Computer Vision -- ECCV 2022 Workshops}, pages 472--484, Cham, 2023. Springer Nature Switzerland.
\newblock ISBN 978-3-031-25082-8.

\bibitem[Kamnitsas et~al.(2017)Kamnitsas, Ledig, Newcombe, Simpson, Kane, Menon, Rueckert, and Glocker]{Kamnitsas2017}
Konstantinos Kamnitsas, Christian Ledig, Virginia~F.J. Newcombe, Joanna~P. Simpson, Andrew~D. Kane, David~K. Menon, Daniel Rueckert, and Ben Glocker.
\newblock Efficient multi-scale 3d cnn with fully connected crf for accurate brain lesion segmentation.
\newblock \emph{Medical Image Analysis}, 36:\penalty0 61--78, 2017.
\newblock ISSN 1361-8415.
\newblock \doi{https://doi.org/10.1016/j.media.2016.10.004}.
\newblock URL \url{https://www.sciencedirect.com/science/article/pii/S1361841516301839}.

\bibitem[Lee(2013)]{Lee2013}
Dong-Hyun Lee.
\newblock Pseudo-label : The simple and efficient semi-supervised learning method for deep neural networks.
\newblock \emph{ICML 2013 Workshop : Challenges in Representation Learning (WREPL)}, 7 2013.

\bibitem[Milletari et~al.(2016)Milletari, Navab, and Ahmadi]{Milletari2016}
Fausto Milletari, Nassir Navab, and Seyed-Ahmad Ahmadi.
\newblock V-net: Fully convolutional neural networks for volumetric medical image segmentation.
\newblock In \emph{2016 Fourth International Conference on 3D Vision (3DV)}, pages 565--571, 2016.
\newblock \doi{10.1109/3DV.2016.79}.

\bibitem[Noroozi and Favaro(2016)]{Noroozi2016}
Mehdi Noroozi and Paolo Favaro.
\newblock Unsupervised learning of visual representations by solving jigsaw puzzles.
\newblock In Bastian Leibe, Jiri Matas, Nicu Sebe, and Max Welling, editors, \emph{Computer Vision -- ECCV 2016}, pages 69--84, Cham, 2016. Springer International Publishing.
\newblock ISBN 978-3-319-46466-4.

\bibitem[O.~Pinheiro et~al.(2020)O.~Pinheiro, Almahairi, Benmalek, Golemo, and Courville]{NEURIPS2020_3000311c}
Pedro~O O.~Pinheiro, Amjad Almahairi, Ryan Benmalek, Florian Golemo, and Aaron~C Courville.
\newblock Unsupervised learning of dense visual representations.
\newblock In H.~Larochelle, M.~Ranzato, R.~Hadsell, M.F. Balcan, and H.~Lin, editors, \emph{Advances in Neural Information Processing Systems}, volume~33, pages 4489--4500. Curran Associates, Inc., 2020.
\newblock URL \url{https://proceedings.neurips.cc/paper_files/paper/2020/file/3000311ca56a1cb93397bc676c0b7fff-Paper.pdf}.

\bibitem[Punn and Agarwal(2022)]{Punn2022}
Narinder~Singh Punn and Sonali Agarwal.
\newblock Bt-unet: A self-supervised learning framework for biomedical image segmentation using barlow twins with u-net models.
\newblock \emph{Machine Learning}, 111:\penalty0 4585--4600, 2022.
\newblock ISSN 1573-0565.
\newblock \doi{10.1007/s10994-022-06219-3}.
\newblock URL \url{https://doi.org/10.1007/s10994-022-06219-3}.

\bibitem[Ronneberger et~al.(2015)Ronneberger, Fischer, and Brox]{Ronneberger2015}
Olaf Ronneberger, Philipp Fischer, and Thomas Brox.
\newblock U-net: Convolutional networks for biomedical image segmentation.
\newblock In Nassir Navab, Joachim Hornegger, William~M. Wells, and Alejandro~F. Frangi, editors, \emph{Medical Image Computing and Computer-Assisted Intervention -- MICCAI 2015}, pages 234--241, Cham, 2015. Springer International Publishing.
\newblock ISBN 978-3-319-24574-4.

\bibitem[Tang et~al.(2022)Tang, Yang, Li, Roth, Landman, Xu, Nath, and Hatamizadeh]{Tang_2022_CVPR}
Yucheng Tang, Dong Yang, Wenqi Li, Holger~R. Roth, Bennett Landman, Daguang Xu, Vishwesh Nath, and Ali Hatamizadeh.
\newblock Self-supervised pre-training of swin transformers for 3d medical image analysis.
\newblock In \emph{Proceedings of the IEEE/CVF Conference on Computer Vision and Pattern Recognition (CVPR)}, pages 20730--20740, June 2022.

\bibitem[Tarvainen and Valpola(2017)]{Tarvainen2017}
Antti Tarvainen and Harri Valpola.
\newblock Mean teachers are better role models: Weight-averaged consistency targets improve semi-supervised deep learning results.
\newblock In I.~Guyon, U.~Von Luxburg, S.~Bengio, H.~Wallach, R.~Fergus, S.~Vishwanathan, and R.~Garnett, editors, \emph{Advances in Neural Information Processing Systems}, volume~30. Curran Associates, Inc., 2017.
\newblock URL \url{https://proceedings.neurips.cc/paper_files/paper/2017/file/68053af2923e00204c3ca7c6a3150cf7-Paper.pdf}.

\bibitem[Tran et~al.(2022)Tran, Wagner, Boxberg, and Peng]{Tran2022}
Manuel Tran, Sophia~J. Wagner, Melanie Boxberg, and Tingying Peng.
\newblock S5cl: Unifying fully-supervised, self-supervised, and semi-supervised learning through hierarchical contrastive learning.
\newblock In Linwei Wang, Qi~Dou, P.~Thomas Fletcher, Stefanie Speidel, and Shuo Li, editors, \emph{Medical Image Computing and Computer Assisted Intervention -- MICCAI 2022}, pages 99--108, Cham, 2022. Springer Nature Switzerland.
\newblock ISBN 978-3-031-16434-7.

\bibitem[Wang et~al.(2021)Wang, Zhang, Shen, Kong, and Li]{Wang_2021_CVPR}
Xinlong Wang, Rufeng Zhang, Chunhua Shen, Tao Kong, and Lei Li.
\newblock Dense contrastive learning for self-supervised visual pre-training.
\newblock In \emph{Proceedings of the IEEE/CVF Conference on Computer Vision and Pattern Recognition (CVPR)}, pages 3024--3033, June 2021.

\bibitem[Xie et~al.(2021)Xie, Lin, Zhang, Cao, Lin, and Hu]{Xie_2021_CVPR}
Zhenda Xie, Yutong Lin, Zheng Zhang, Yue Cao, Stephen Lin, and Han Hu.
\newblock Propagate yourself: Exploring pixel-level consistency for unsupervised visual representation learning.
\newblock In \emph{Proceedings of the IEEE/CVF Conference on Computer Vision and Pattern Recognition (CVPR)}, pages 16684--16693, June 2021.

\bibitem[Yan et~al.(2022)Yan, Cai, Jin, Miao, Guo, Harrison, Tang, Xiao, Lu, and Lu]{Ke2022}
Ke~Yan, Jinzheng Cai, Dakai Jin, Shun Miao, Dazhou Guo, Adam~P. Harrison, Youbao Tang, Jing Xiao, Jingjing Lu, and Le~Lu.
\newblock Sam: Self-supervised learning of pixel-wise anatomical embeddings in radiological images.
\newblock \emph{IEEE Transactions on Medical Imaging}, 41\penalty0 (10):\penalty0 2658--2669, 2022.
\newblock \doi{10.1109/TMI.2022.3169003}.

\bibitem[Yu et~al.(2019)Yu, Wang, Li, Fu, and Heng]{Yu2019}
Lequan Yu, Shujun Wang, Xiaomeng Li, Chi-Wing Fu, and Pheng-Ann Heng.
\newblock Uncertainty-aware self-ensembling model for semi-supervised 3d left atrium segmentation.
\newblock In Dinggang Shen, Tianming Liu, Terry~M. Peters, Lawrence~H. Staib, Caroline Essert, Sean Zhou, Pew-Thian Yap, and Ali Khan, editors, \emph{Medical Image Computing and Computer Assisted Intervention -- MICCAI 2019}, pages 605--613, Cham, 2019. Springer International Publishing.
\newblock ISBN 978-3-030-32245-8.

\bibitem[Zbontar et~al.(2021)Zbontar, Jing, Misra, LeCun, and Deny]{zbontar21a}
Jure Zbontar, Li~Jing, Ishan Misra, Yann LeCun, and Stephane Deny.
\newblock Barlow twins: Self-supervised learning via redundancy reduction.
\newblock In Marina Meila and Tong Zhang, editors, \emph{Proceedings of the 38th International Conference on Machine Learning}, volume 139 of \emph{Proceedings of Machine Learning Research}, pages 12310--12320. PMLR, 18--24 Jul 2021.
\newblock URL \url{https://proceedings.mlr.press/v139/zbontar21a.html}.

\bibitem[Zeng et~al.(2021)Zeng, Wu, Hu, Xu, Yuan, Huang, Zhuang, Hu, and Shi]{Zeng2021}
Dewen Zeng, Yawen Wu, Xinrong Hu, Xiaowei Xu, Haiyun Yuan, Meiping Huang, Jian Zhuang, Jingtong Hu, and Yiyu Shi.
\newblock Positional contrastive learning for volumetric medical image segmentation.
\newblock In Marleen de~Bruijne, Philippe~C. Cattin, St{\'e}phane Cotin, Nicolas Padoy, Stefanie Speidel, Yefeng Zheng, and Caroline Essert, editors, \emph{Medical Image Computing and Computer Assisted Intervention -- MICCAI 2021}, pages 221--230, Cham, 2021. Springer International Publishing.
\newblock ISBN 978-3-030-87196-3.

\bibitem[Zhang et~al.(2016)Zhang, Isola, and Efros]{Zhang2016colorful}
Richard Zhang, Phillip Isola, and Alexei~A. Efros.
\newblock Colorful image colorization.
\newblock In Bastian Leibe, Jiri Matas, Nicu Sebe, and Max Welling, editors, \emph{Computer Vision -- ECCV 2016}, pages 649--666, Cham, 2016. Springer International Publishing.
\newblock ISBN 978-3-319-46487-9.

\bibitem[Zhao et~al.(2021)Zhao, Vemulapalli, Mansfield, Gong, Green, Shapira, and Wu]{Zhao_2021_ICCV}
Xiangyun Zhao, Raviteja Vemulapalli, Philip~Andrew Mansfield, Boqing Gong, Bradley Green, Lior Shapira, and Ying Wu.
\newblock Contrastive learning for label efficient semantic segmentation.
\newblock In \emph{Proceedings of the IEEE/CVF International Conference on Computer Vision (ICCV)}, pages 10623--10633, October 2021.

\bibitem[Zhong et~al.(2021)Zhong, Yuan, Wu, Yuan, Peng, and Wang]{Zhong_2021_ICCV}
Yuanyi Zhong, Bodi Yuan, Hong Wu, Zhiqiang Yuan, Jian Peng, and Yu-Xiong Wang.
\newblock Pixel contrastive-consistent semi-supervised semantic segmentation.
\newblock In \emph{Proceedings of the IEEE/CVF International Conference on Computer Vision (ICCV)}, pages 7273--7282, October 2021.

\end{thebibliography}

\newpage
\appendix

\section{Additional Results}

\noindent\textbf{Pix2Rep-v2 + Mean Teacher.} As a combined method, we have also tested the combination of pretraining via \texttt{Pix2Rep-v2} with semi-supervised fine-tuning (Mean Teacher). Results are provided in \tableref{tab: BT - MT}:

\begin{table}[h]
    \centering
    \begin{tabular}{p{0.35\textwidth} p{0.05\textwidth} p{0.05\textwidth} p{0.05\textwidth} p{0.05\textwidth} p{0.05\textwidth} p{0.05\textwidth} p{0.05\textwidth}}
    \bottomrule
        \textbf{Method:\hspace{0.5cm}~/\hspace{0.5cm}~$|X_{tr}|\coloneqq$} & $1$ & $2$ & $5$ & $10$ & $20$ & $50$ & $100$ \\
    \bottomrule
        \texttt{Pix2Rep-v2} + Mean Teacher & $\mathbf{87.9}$ & $\mathbf{90.0}$ & $\mathbf{{91.8}}$ & $\mathbf{{92.4}}$ & $\mathbf{{92.7}}$ & ${93.4}$ & ${93.9}$ \\
    \bottomrule
    \end{tabular}
    \caption{Performance on the ACDC dataset of \texttt{Pix2Rep-v2} followed by (Mean Teacher) semi-supervised fine-tuning. Results are highlighted in \textbf{bold} when above the results reported in \tableref{tab: benchmark}.
    }
    \label{tab: BT - MT}
\end{table}

\section{Dice Scores per Anatomical Structures}

Table~\ref{tab: benchmark} reports the  test Dice scores averaged over the four segmented structures: Left Ventricle, Right Ventricle, Myocardium and Background. We report the corresponding Dice scores over individual anatomical structures in Tables~\ref{tab: LV Dice}, \ref{tab: RV Dice}, \ref{tab: MYO Dice}, \ref{tab: BG Dice}.

\begin{table}[h]
    \centering
    \begin{tabular}{p{0.42\textwidth} p{0.05\textwidth} p{0.05\textwidth} p{0.05\textwidth} p{0.05\textwidth} p{0.05\textwidth} p{0.05\textwidth} p{0.05\textwidth}}
            & \multicolumn{7}{c}{\cellcolor{LightCyan}\textbf{LV~Dice (ACDC), for $|X_{tr}|$ set to:}} \\
    \bottomrule
        \multicolumn{8}{c}{\vspace{-0.85em}} \\
        \textbf{Method:\hspace{1.0cm}~/\hspace{1.0cm}~$|X_{tr}|\coloneqq$} & $1$ & $2$ & $5$ & $10$ & $20$ & $50$ & $100$ \\
    \bottomrule
        \rowcolor{Gray}
        \multicolumn{8}{c}{\textbf{Fully-supervised learning:}} \\
        \multicolumn{8}{c}{\vspace{-0.95em}} \\
        Baseline (U-Net) & $49.2$ & $70.2$ & $82.7$ & $87.4$ & $93.7$ & $95.0$ & $95.6$ \\
        \rowcolor{Gray}
        \multicolumn{8}{c}{\textbf{Semi-supervised learning:}} \\
        \multicolumn{8}{c}{\vspace{-0.95em}} \\
        Mean Teacher & $59.7$ & $75.1$ & $89.9$ & $93.1$ & $94.1$ & $95.5$ & $\mathbf{96.4}$ \\
        \rowcolor{Gray}
        \multicolumn{8}{c}{\textbf{Self-supervised learning (+ Fine-tuning by default):}} \\
        \multicolumn{8}{c}{\vspace{-0.95em}} \\
        SimCLR pretraining & $66.9$ & $83.9$ & $90.5$ & $92.8$ & $93.8$ & $95.2$ & $95.3$ \\
        Chaitanya et al. (2020) & $82.0$ & $80.2$ & $89.5$ & $91.6$ & $94.8$ & $95.7$ & $96.1$ \\
        \textbf{Proposed} (w/o rot. \& int. reversal) & $84.1$ & $84.2$ & $92.2$ & $93.9$ & $95.0$ & $95.8$ & ${96.2}$ \\
        \textbf{Proposed} (only Linear-probing) & $88.6$ & $90.2$ & $91.3$ & $92.0$ & $92.5$ & $93.1$ & $93.4$ \\
        \textbf{Proposed} & \underline{$89.9$} & \underline{$92.7$} & \underline{$93.9$} & \underline{$94.3$} & \underline{$95.3$} & \underline{$95.9$} & \underline{$96.3$} \\
        \rowcolor{Gray}
        \multicolumn{8}{c}{\textbf{Combined method:}} \\
        \multicolumn{8}{c}{\vspace{-0.95em}} \\
        \textbf{Proposed} + Mean Teacher & $\mathbf{91.2}$ & $\mathbf{93.2}$ & $\mathbf{94.3}$ & $\mathbf{95.3}$ & $\mathbf{95.7}$ & $\mathbf{96.3}$ & {$96.1$} \\
    \bottomrule
    \end{tabular}
    \caption{Comparison with the state-of-the-art on the test ACDC dataset, for various amounts of labeled training data $|X_{tr}|$, in terms of Left Ventricle (LV) Dice overlap. Each score is averaged over five runs (see main text -- Evaluation).}
    \label{tab: LV Dice}
\end{table}

\begin{table}[h]
    \centering
    \begin{tabular}{p{0.42\textwidth} p{0.05\textwidth} p{0.05\textwidth} p{0.05\textwidth} p{0.05\textwidth} p{0.05\textwidth} p{0.05\textwidth} p{0.05\textwidth}}
            & \multicolumn{7}{c}{\cellcolor{LightCyan}\textbf{RV~Dice (ACDC), for $|X_{tr}|$ set to:}} \\
    \bottomrule
        \multicolumn{8}{c}{\vspace{-0.85em}} \\
        \textbf{Method:\hspace{1.0cm}~/\hspace{1.0cm}~$|X_{tr}|\coloneqq$} & $1$ & $2$ & $5$ & $10$ & $20$ & $50$ & $100$ \\
    \bottomrule
        \rowcolor{Gray}
        \multicolumn{8}{c}{\textbf{Fully-supervised learning:}} \\
        \multicolumn{8}{c}{\vspace{-0.95em}} \\
        Baseline (U-Net) & $41.1$ & $54.9$ & $69.8$ & $74.5$ & $84.8$ & $88.5$ & $88.8$ \\
        \rowcolor{Gray}
        \multicolumn{8}{c}{\textbf{Semi-supervised learning:}} \\
        \multicolumn{8}{c}{\vspace{-0.95em}} \\
        Mean Teacher & $40.9$ & $53.1$ & $74.1$ & $81.2$ & $82.8$ & $86.7$ & $89.3$ \\
        \rowcolor{Gray}
        \multicolumn{8}{c}{\textbf{Self-supervised learning (+ Fine-tuning by default):}} \\
        \multicolumn{8}{c}{\vspace{-0.95em}} \\
        SimCLR pretraining & $43.4$ & $68.0$ & $75.2$ & $81.2$ & $83.1$ & $86.2$ & $87.2$ \\
        Chaitanya et al. (2020) & $60.4$ & $61.4$ & $79.7$ & $82.1$ & $83.9$ & $86.6$ & $87.3$ \\
        \textbf{Proposed} (w/o rot. \& int. reversal) & $57.9$ & $69.6$ & $80.3$ & $83.8$ & $87.2$ & $89.2$ & $\mathbf{91.5}$ \\
        \textbf{Proposed} (only Linear-probing) & $71.0$ & $74.7$ & $77.2$ & $78.4$ & $80.1$ & $81.2$ & $81.8$ \\
        \textbf{Proposed} & \underline{$74.0$} & \underline{$79.1$} & \underline{$83.0$} & \underline{$85.2$} & {$\mathbf{87.7}$} & {$\mathbf{89.7}$} & \underline{${90.5}$} \\
        \rowcolor{Gray}
        \multicolumn{8}{c}{\textbf{Combined method:}} \\
        \multicolumn{8}{c}{\vspace{-0.95em}} \\
        \textbf{Proposed} + Mean Teacher & $\mathbf{75.2}$ & $\mathbf{81.9}$ & $\mathbf{84.6}$ & $\mathbf{85.9}$ & \underline{${87.5}$} & \underline{${89.4}$} & {$90.4$} \\
    \bottomrule
    \end{tabular}
    \caption{Comparison with the state-of-the-art on the test ACDC dataset, for various amounts of labeled training data $|X_{tr}|$, in terms of Right Ventricle (RV) Dice overlap. Each score is averaged over five runs (see main text -- Evaluation).}
    \label{tab: RV Dice}
\end{table}

\begin{table}[h!]
    \centering
    \begin{tabular}{p{0.42\textwidth} p{0.05\textwidth} p{0.05\textwidth} p{0.05\textwidth} p{0.05\textwidth} p{0.05\textwidth} p{0.05\textwidth} p{0.05\textwidth}}
            & \multicolumn{7}{c}{\cellcolor{LightCyan}\textbf{MYO~Dice (ACDC), for $|X_{tr}|$ set to:}} \\
    \bottomrule
        \multicolumn{8}{c}{\vspace{-0.85em}} \\
        \textbf{Method:\hspace{1.0cm}~/\hspace{1.0cm}~$|X_{tr}|\coloneqq$} & $1$ & $2$ & $5$ & $10$ & $20$ & $50$ & $100$ \\
    \bottomrule
        \rowcolor{Gray}
        \multicolumn{8}{c}{\textbf{Fully-supervised learning:}} \\
        \multicolumn{8}{c}{\vspace{-0.95em}} \\
        Baseline (U-Net) & $34.7$ & $52.1$ & $69.2$ & $74.8$ & $86.3$ & $89.3$ & $89.9$ \\
        \rowcolor{Gray}
        \multicolumn{8}{c}{\textbf{Semi-supervised learning:}} \\
        \multicolumn{8}{c}{\vspace{-0.95em}} \\
        Mean Teacher & $46.0$ & $59.0$ & $78.1$ & $84.3$ & $86.2$ & $89.4$ & \underline{$91.3$} \\
        \rowcolor{Gray}
        \multicolumn{8}{c}{\textbf{Self-supervised learning (+ Fine-tuning by default):}} \\
        \multicolumn{8}{c}{\vspace{-0.95em}} \\
        SimCLR pretraining & $53.1$ & $64.5$ & $77.8$ & $84.4$ & $86.5$ & $88.9$ & $89.8$ \\
        Chaitanya et al. (2020) & $65.4$ & $71.0$ & $74.8$ & $81.7$ & $84.7$ & $87.1$ & $87.8$ \\
        \textbf{Proposed} (w/o rot. \& int. reversal) & $74.7$ & $73.0$ & $82.1$ & $85.6$ & $88.6$ & $90.1$ & ${90.9}$ \\
        \textbf{Proposed} (only Linear-probing) & $77.4$ & $81.2$ & $82.5$ & $83.2$ & $84.6$ & $85.6$ & $85.9$ \\
        \textbf{Proposed} & \underline{$78.3$} & \underline{$83.7$} & \underline{$86.3$} & \underline{$87.5$} & \underline{$88.9$} & \underline{$90.4$} & $91.0$ \\
        \rowcolor{Gray}
        \multicolumn{8}{c}{\textbf{Combined method:}} \\
        \multicolumn{8}{c}{\vspace{-0.95em}} \\
        \textbf{Proposed} + Mean Teacher & $\mathbf{81.4}$ & $\mathbf{87.0}$ & $\mathbf{87.9}$ & $\mathbf{88.8}$ & $\mathbf{89.4}$ & $\mathbf{91.1}$ & {$\mathbf{91.9}$} \\
    \bottomrule
    \end{tabular}
    \caption{Comparison with the state-of-the-art on the test ACDC dataset, for various amounts of labeled training data $|X_{tr}|$, in terms of Myocardium (MYO) Dice overlap. Each score is averaged over five runs (see main text -- Evaluation).}
    \label{tab: MYO Dice}
\end{table}

\begin{table}[h!]
    \centering
    \begin{tabular}{p{0.42\textwidth} p{0.05\textwidth} p{0.05\textwidth} p{0.05\textwidth} p{0.05\textwidth} p{0.05\textwidth} p{0.05\textwidth} p{0.05\textwidth}}
            & \multicolumn{7}{c}{\cellcolor{LightCyan}\textbf{BG~Dice (ACDC), for $|X_{tr}|$ set to:}} \\
    \bottomrule
        \multicolumn{8}{c}{\vspace{-0.85em}} \\
        \textbf{Method:\hspace{1.0cm}~/\hspace{1.0cm}~$|X_{tr}|\coloneqq$} & $1$ & $2$ & $5$ & $10$ & $20$ & $50$ & $100$ \\
    \bottomrule
        \rowcolor{Gray}
        \multicolumn{8}{c}{\textbf{Fully-supervised learning:}} \\
        \multicolumn{8}{c}{\vspace{-0.95em}} \\
        Baseline (U-Net) & $89.2$ & $92.0$ & $95.3$ & $95.9$ & $97.7$ & $98.3$ & $98.3$ \\
        \rowcolor{Gray}
        \multicolumn{8}{c}{\textbf{Semi-supervised learning:}} \\
        \multicolumn{8}{c}{\vspace{-0.95em}} \\
        Mean Teacher & $82.5$ & $88.3$ & $95.9$ & $97.2$ & $97.4$ & $98.1$ & $98.5$ \\
        \rowcolor{Gray}
        \multicolumn{8}{c}{\textbf{Self-supervised learning (+ Fine-tuning by default):}} \\
        \multicolumn{8}{c}{\vspace{-0.95em}} \\
        SimCLR pretraining & $89.7$ & $94.9$ & $96.4$ & $97.3$ & $97.5$ & $98.0$ & $98.2$ \\
        Chaitanya et al. (2020) & $\mathbf{99.2}$ & $\mathbf{99.3}$ & $\mathbf{99.5}$ & $\mathbf{99.6}$ & $\mathbf{99.7}$ & $\mathbf{99.7}$ & $\mathbf{99.7}$ \\
        \textbf{Proposed} (w/o rot. \& int. reversal) & $94.3$ & $95.1$ & $96.8$ & $97.5$ & $98.1$ & $98.4$ & \underline{${98.6}$} \\
        \textbf{Proposed} (only Linear-probing) & $96.2$ & $96.7$ & $97.1$ & $97.1$ & $97.3$ & $97.5$ & $97.5$ \\
        \textbf{Proposed} & \underline{$96.6$} & \underline{$97.2$} & \underline{$97.6$} & \underline{$97.9$} & \underline{$98.2$} & \underline{$98.5$} & \underline{$98.6$} \\
        \rowcolor{Gray}
        \multicolumn{8}{c}{\textbf{Combined method:}} \\
        \multicolumn{8}{c}{\vspace{-0.95em}} \\
        \textbf{Proposed} + Mean Teacher & \underline{${96.6}$} & ${96.9}$ & ${97.5}$ & ${97.8}$ & ${98.1}$ & ${98.3}$ & {$98.3$} \\
    \bottomrule
    \end{tabular}
    \caption{Comparison with the state-of-the-art on the test ACDC dataset, for various amounts of labeled training data $|X_{tr}|$, in terms of Dice overlap for the background (BG). Each score is averaged over five runs (see main text -- Evaluation).}
    \label{tab: BG Dice}
\end{table}

\section{Visualization of the \texttt{Pix2Rep} Pixel Embeddings}

To gain qualitative insights into the learned pixel representations, Fig.~\ref{fig: pixel embeddings visualization} graphically presents pixel embeddings returned by our pretrained \texttt{Pix2Rep} model, learned  without supervision and \textit{prior} to the fine-tuning stage of the downstream segmentation task. 

Firstly, we show 2D \texttt{t-SNE} projections of all pixel embeddings for three  test images, color-coded by their true class (background, left ventricle, right ventricle, myocardium). We can see that the four anatomical structures are well separated in the representation space, despite not using any label supervision.

Secondly, we assign to each 2D \texttt{t-SNE} coordinate positions a color, according to a reference colormap shown at the top right of Fig.~\ref{fig: pixel embeddings visualization}. Then, we color-code each pixel of each test image by its color as obtained by this scheme (MRI image pixel $\mapsto$ pixel \texttt{Pix2Rep} embedding $\mapsto$ pixel embedding mapped to 2D coordinates after  \texttt{t-SNE} projection $\mapsto$ pixel embedding mapped to an individual color value with a 2D reference colormap applied on the 2D \texttt{t-SNE} space $\mapsto$ colored pixel embedding displayed in original MRI image space). We remark that visually, our \texttt{Pix2Rep} embedding leads to coarse segmentations of the cardiac structures, again without involving any supervision with label annotations.

\begin{figure}[t!]
\floatconts
  {fig: pixel embeddings visualization}
  {\caption{\texttt{Pix2Rep} pixel-level embeddings. First and Second columns: test cardiac MRI images and ground truth segmentations. Third column: 2D \texttt{t-SNE} coordinates of \texttt{Pix2Rep} pixel embeddings. Fourth column: colored pixel embedding displayed in original MRI image space. The reference colormap used to map 2D \texttt{t-SNE} coordinates with individual colors is shown in the vignette on the top row example.
  }}
  {\includegraphics[width=1\textwidth]{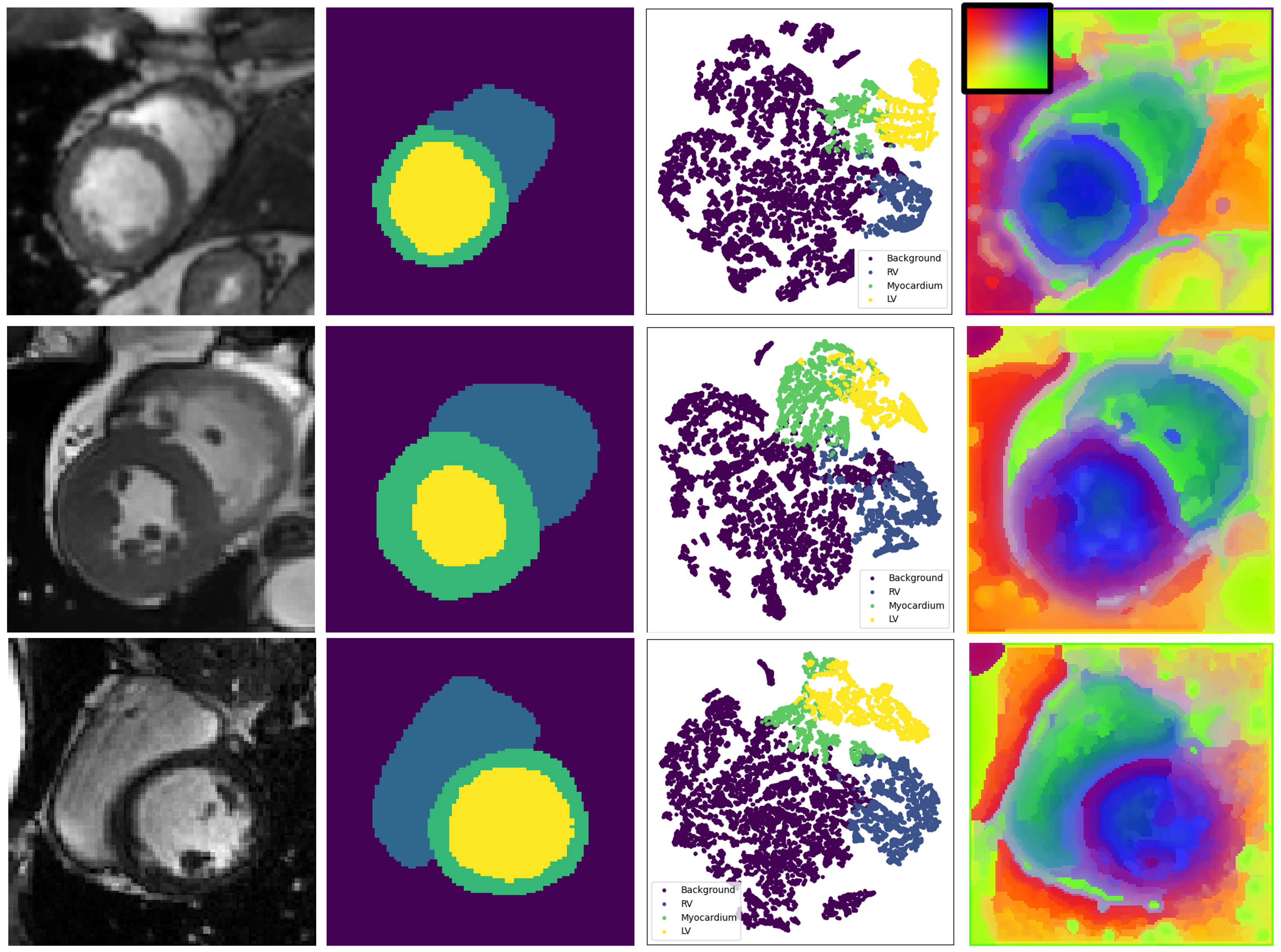}}
\end{figure}

\end{document}